\documentclass[10pt,twocolumn,letterpaper]{article}

\usepackage[pagenumbers]{cvpr}

\usepackage{indentfirst}
\usepackage{float}

\definecolor{cvprblue}{rgb}{0.21,0.49,0.74}
\usepackage[pagebackref,breaklinks,colorlinks,allcolors=cvprblue]{hyperref}
\hypersetup{
  pdftitle={VLX-VR: An Agentic-Aware Video Reasoning Model},
  pdfauthor={Sheng Li, Peng Liu, Qianqian Zhang, Tiancheng Zhao},
  pdfsubject={VLX-VR Technical Report},
  pdfkeywords={VLX-VR, video reasoning, multimodal memory, agentic-aware vision-language model}
}

\title{VLX-VR: An Agentic-Aware Video Reasoning Model}

\author{Sheng Li \quad Peng Liu \quad Qianqian Zhang \quad Tiancheng Zhao \\
Om AI Research\\ 
{\tt\small tianchez@zju-bj.com}\\
\small Correspondence: Tiancheng Zhao 
}

\begin{document}
\maketitle
\begin{abstract}
Real-world video understanding requires integrating visual, audio, textual, and temporal evidence distributed across a video. Yet many pipelines use a fixed video context and single-pass inference, limiting adaptive evidence acquisition when observations are incomplete, ambiguous, or conflicting. We present \textbf{VLX-VR}, an agentic-aware video reasoning model trained within a video reasoning framework defined by a Think--Memory--Observation loop. At each step, VLX-VR determines the needed evidence, invokes \texttt{read\_memory} or \texttt{write\_memory}, incorporates the returned Observation, and decides whether to continue or produce the task output. We train VLX-VR with multimodal data, including videos and agent trajectories, using reinforcement learning to learn evidence acquisition, memory use, and termination. On MINERVA, VLX-VR achieves state-of-the-art performance among the models included in our comparison, with 78.79\% accuracy. Under the original three duration groups, its accuracies are 76.70\%, 78.73\%, and 80.92\%, with a cross-duration accuracy variance of 2.97~$\mathrm{pp}^2$. On correctly answered samples, 96.20\% of VLX-VR's reasoning traces are consistent with the MINERVA reference reasoning traces and the evidence described by them, while approximately 75.80\% of all evaluated samples satisfy both answer correctness and this evidence-grounded trace criterion. These results show strong performance and broadly stable behavior across durations, while counting, state changes, causal reasoning, and spatial perception remain challenging.
\par\medskip
\end{abstract}
    
\section{Introduction}
\label{sec:intro}

In everyday scenarios, people understand videos by combining visual, audio, textual, and temporal evidence to analyze events and make judgments. Determining what happened, why it happened, or how a situation changed may require locating relevant moments, comparing states before and after an event, reading text in a scene, and relating spoken content to visual actions. Supporting such real-world reasoning requires more than recognizing objects or isolated frames: a model must identify, integrate, and reassess evidence distributed across the video before reaching a reliable conclusion.

In many standard video-analysis and VideoQA pipelines, video large language models such as Video-LLaVA~\cite{videollava}, Qwen3-VL~\cite{team2025qwen3}, and VideoLLaMA 3~\cite{videollama3} typically receive sampled frames or video clips as a fixed input and produce an output in a single inference pass. This fixed-input, single-pass pattern cannot fully satisfy the real-world event-analysis requirements described above. In such scenarios, a model may need to revisit an earlier event, seek additional evidence after an initial observation, or compare information distributed across time and modalities before making a judgment. Because the evidence is determined before reasoning begins, the model cannot adaptively acquire new evidence after detecting missing, ambiguous, or conflicting information. As a result, insufficient observation may omit a decisive event, whereas indiscriminate observation increases context length and inference cost.

Existing video agents move closer to these real-world requirements through iterative information gathering, temporal localization, and memory queries~\cite{videoagent_wang,videoagent_fan,videoagent2}. However, they may still fall short when a general-purpose VLM is merely placed inside an external agent loop without being trained for its role in that loop. In this setting, the model has not necessarily learned to treat evidence acquisition, memory use, and termination as parts of its reasoning policy. It may request redundant evidence, fail to retain important intermediate states, or produce a result before the available evidence is sufficient. Therefore, reliably analyzing events and making judgments in real-world videos still requires stronger temporal localization, multimodal evidence selection, state maintenance, and consistency between reasoning and actual observations~\cite{minerva}.

These limitations motivate the following question: \emph{Can a video reasoning model be trained not merely to participate in an agentic pipeline, but to learn evidence acquisition, memory use, and termination as integral parts of its reasoning process?}

To address this question, we introduce a video reasoning framework organized as a Think--Memory--Observation loop and present VLX-VR, an agentic-aware video reasoning model trained to operate within this framework. VLX-VR targets video-analysis tasks that require temporal localization, multimodal evidence selection, state tracking, and multi-step reasoning over both short and long videos. Instead of relying on a fixed video context, VLX-VR learns to acquire and reassess multimodal evidence through the framework-defined loop with direct access to multimodal memory. We train VLX-VR with multimodal data and agent trajectories using reinforcement learning so that evidence acquisition, memory use, and termination become parts of its learned reasoning policy. Section~\ref{sec:method} describes the framework and training procedure in detail.

\textbf{Our contributions are:}
\begin{enumerate}
  \item We introduce a video reasoning framework that organizes task-conditioned evidence acquisition as a Think--Memory--Observation loop with direct multimodal-memory access.
  \item We train VLX-VR as an agentic-aware video reasoning model with multimodal data and agent trajectories using reinforcement learning, enabling the model to learn evidence acquisition, memory read/write behavior, Observation-based reasoning, and termination within the framework-defined loop.
  \item VLX-VR achieves state-of-the-art performance on MINERVA, reaching 78.79\% accuracy among the models included in our comparison. We further evaluate its cross-duration stability and the evidence-grounded agreement between its reasoning traces and the MINERVA reference reasoning traces.
\end{enumerate}

\section{Related Work}
\label{sec:related}

\subsection{General Video Understanding Models}
Vision-language models (VLMs) extend language models with visual perception and cross-modal alignment, providing the foundation for multimodal video understanding~\cite{song2024moviechat}. Video-LLaVA learns a shared representation space for images and videos~\cite{videollava}. Qwen2.5-VL advances visual understanding, temporal localization, and long-video processing~\cite{qwen25vl}, while VideoLLaMA 3 develops a vision-centric training strategy for general image and video understanding~\cite{videollama3}. These models support tasks such as video description, dialogue, event interpretation, temporal analysis, and VideoQA. Within this line of research, OmChat~\cite{omchat2024} focuses on native multimodal modeling with strong long-context and video-understanding capabilities. It combines dynamic visual encoding with progressive multimodal pretraining to process images, multiple images, and long videos in a unified model. VLX-Flow extends model-level video understanding from offline clips to continuous video streams. It incrementally maintains visual context and semantic memory, allowing previously observed content to be reused without reprocessing the complete video history~\cite{vlxflow2026}.

These studies strengthen the perceptual, alignment, temporal-modeling, and context-maintenance capabilities required for video understanding. VLX-VR builds on these foundations but focuses on a broader control problem: learning how to acquire task-relevant evidence, retain useful intermediate information, and determine whether the available evidence is sufficient to complete the current video-analysis task.

\subsection{From Perception to Visual Reasoning}
Beyond perception and cross-modal alignment, recent research has increasingly focused on enabling models to perform explicit and multi-step reasoning~\cite{xie2025video,zhang2026thinking}. Chain-of-thought prompting demonstrates that generating intermediate reasoning steps can improve performance on complex language tasks~\cite{wei2022cot,zhang2025r1,yang2025wethink}. Reasoning models such as OpenAI o1 and DeepSeek-R1 further show that reinforcement learning can train models to refine intermediate reasoning strategies before producing a final response~\cite{openai2024o1,deepseekr1}.

This direction has subsequently been extended to visual reasoning. VLM-R1 applies rule-based reinforcement learning to general vision-language tasks and studies the stability and generalization of R1-style training for VLMs~\cite{vlmr1_2025}. Video-R1 introduces temporal-aware reinforcement learning for video reasoning and combines image and video reasoning data during training~\cite{videor1}. Together, these studies mark a transition from recognizing visual content to deriving conclusions through structured reasoning over visual and temporal evidence.

VLX-VR follows this transition but extends the learning target beyond generating an intermediate reasoning trace. Its reasoning process also controls multimodal-memory reads and writes, incorporates the Observations returned by these operations, and determines whether additional evidence is required before producing the task output. Reasoning therefore serves not only to explain a result, but also to control how the model observes and analyzes the video.

\subsection{Multimodal Agents for Visual Analysis}
A complementary line of research combines multimodal models with memory, tools, and iterative interaction~\cite{wang2024videoagent,ma2025drvideo}. ReAct establishes the reasoning--action--observation paradigm, in which intermediate results guide subsequent reasoning and actions~\cite{react}. For complex video understanding, OmAgent introduces a multimodal agent framework that stores and retrieves relevant video frames and uses a task divide-and-conquer loop to dynamically invoke tools and process complex video-analysis requests~\cite{omagent2024}.

Related video-agent systems explore different mechanisms for adaptive evidence acquisition~\cite{liu2026videomind,yeo2026worldmm}. VideoAgent by Wang et al. iteratively identifies and collects task-relevant visual information from long-form videos using visual tools~\cite{videoagent_wang}. VideoAgent by Fan et al. constructs structured event and object memory and uses temporal localization and memory-query tools~\cite{videoagent_fan}. VideoAgent2 further employs uncertainty-aware reasoning to regulate information gathering~\cite{videoagent2}. These studies demonstrate that iterative observation, external memory, and tool use can reduce the limitations of analyzing a fixed and preselected video context.

Building on these agent-based approaches, VLX-VR is trained as an agentic-aware VLM rather than relying solely on external inference-time control. Within the framework-defined Think--Memory--Observation loop, VLX-VR conditions its decisions on the task instruction, current reasoning state, accumulated multimodal memory, and returned Observations. Memory directly provides the \texttt{read\_memory} and \texttt{write\_memory} operations; VLX-VR selects one of them, integrates the resulting evidence or update status, and decides whether to continue the loop or produce the task output. The central objective is to make interaction with multimodal memory part of the model's learned reasoning policy.

\section{The Proposed Method}
\label{sec:method}

\subsection{Framework Overview}
\begin{figure*}[t]
  \centering
  \includegraphics[width=\textwidth]{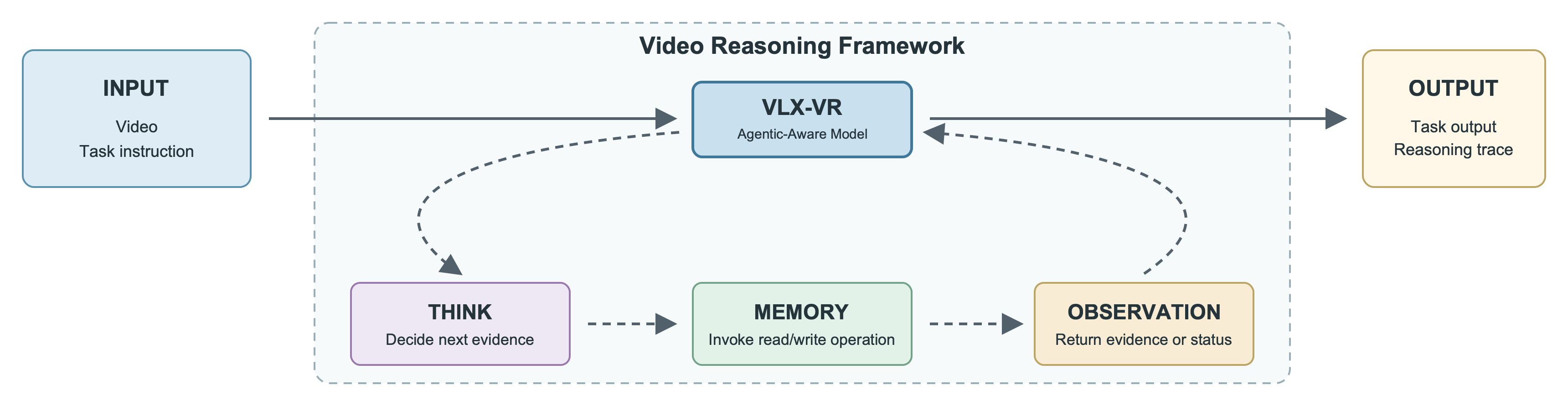}
  \caption{Overview of the framework-defined Think--Memory--Observation loop used to train and run VLX-VR, an agentic-aware video reasoning model. Memory directly provides \texttt{read\_memory} and \texttt{write\_memory}; VLX-VR invokes one of these operations, and the resulting Observation guides the next Think step or the final output.}
  \label{fig:architecture}
\end{figure*}

Given a video and a task instruction, the agentic-aware VLX-VR model operates within the framework-defined Think--Memory--Observation loop. VLX-VR first performs Think to determine what evidence is needed. It then enters Memory and directly invokes the provided \texttt{read\_memory} or \texttt{write\_memory} operation to retrieve relevant multimodal evidence or retain intermediate state. The result is returned as an Observation, which VLX-VR uses to update its reasoning state. VLX-VR then decides whether to gather additional evidence or produce a final result. As illustrated in Figure~\ref{fig:architecture}, evidence acquisition is conditioned on the task objective, accumulated memory, and returned Observations rather than fixed before reasoning begins.

\subsection{Think--Memory--Observation loop}
At reasoning step $t$, VLX-VR maintains a state $s_t$ that includes the task instruction, observed evidence, multimodal content stored in memory, current output hypotheses, and unresolved uncertainty. VLX-VR generates Think from $s_t$ and decides whether to invoke \texttt{read\_memory} or \texttt{write\_memory} during Memory, or terminate. This state-dependent decision allows the evidence request to change as new Observations become available.

The reasoning loop consists of three stages:
\begin{itemize}
  \item \textbf{Think:} VLX-VR interprets the task objective, proposes the next evidence need, and estimates whether the current evidence is sufficient.
  \item \textbf{Memory:} Memory directly provides \texttt{read\_memory} and \texttt{write\_memory}; VLX-VR invokes one operation to retrieve or retain multimodal evidence.
  \item \textbf{Observation:} The result of the selected Memory operation is returned to VLX-VR for the next decision.
\end{itemize}
When evidence is sufficient, VLX-VR generates the final output with supporting evidence. When evidence is insufficient or conflicting, the model returns to Think and starts another Memory step. The minimal operation space avoids dependence on a large collection of specialized tools and makes training and evaluation easier to standardize.

The multimodal memory serves as an external and inspectable reasoning state rather than only a cache of textual summaries. Video, audio, supporting evidence, Observations, and intermediate states can be read, written, recorded, and reused across reasoning steps. This design also makes it possible to analyze whether the final output is supported by the evidence actually observed during inference.

Evidence and reasoning traces play different but related roles in VLX-VR. Evidence refers to task-relevant multimodal content acquired from the video or returned by Memory, including visual regions, audio segments, textual cues, timestamps, and intermediate state information. A reasoning trace is the ordered record of how VLX-VR interprets the task, selects Memory operations, incorporates returned Observations, connects evidence across steps, and reaches the final output. Thus, evidence provides the basis for the reasoning trace, while the reasoning trace determines which evidence to acquire, how to organize and interpret it, and whether the available evidence is sufficient for the task. Agreement between a reasoning trace and its supporting evidence indicates that the stated reasoning is grounded in the observed information; it does not by itself establish that the trace faithfully reveals the model's internal decision process.

The Memory stage provides two operations directly to VLX-VR. \texttt{read\_memory} retrieves video, audio, and other multimodal content relevant to the current task objective. \texttt{write\_memory} stores the current Observation, evidence, and state information in multimodal memory. Each operation returns an Observation containing either the requested evidence or the status of the memory update.

The inference process can be written as:
\begin{verbatim}
s_0 = initialize(video, instruction)
while not stop(s_t):
    think_t, call_t = VLX-VR(s_t)
    obs_t = execute(call_t)
    s_{t+1} = update(s_t, obs_t)
output = VLX-VR(s_t)
\end{verbatim}
Here, \texttt{call\_t} is either \texttt{read\_memory} or \texttt{write\_memory}, selected directly by VLX-VR from the current reasoning state, and \texttt{obs\_t} is the resulting Observation. The stopping condition depends on evidence sufficiency, output confidence, loop budget, and failure state. Premature termination with insufficient evidence should receive a lower reward. Repeatedly reading the same content or extending the loop without a useful objective should be penalized through cost and redundancy terms.

\subsection{Training}
Reinforcement learning has been used to incentivize language-model reasoning in DeepSeek-R1~\cite{deepseekr1}. For visual tasks, VLM-R1 investigates rule-based reinforcement learning for visual grounding and open-vocabulary object detection, reporting improved generalization over supervised fine-tuning in the evaluated settings~\cite{vlmr1_2025}. Video-R1 studies video reasoning through temporal-aware T-GRPO optimization and mixed image--video training~\cite{videor1}. Search-R1 provides a complementary text-domain example of learning interleaved reasoning and search-tool interactions with reinforcement learning~\cite{searchr1}. These works motivate reinforcement learning for visual tasks and tool interaction, without implying that they implement our Think--Memory--Observation loop.

We train VLX-VR as an agentic-aware video reasoning model within this framework rather than only connecting a general-purpose model to an externally defined loop at inference time. Multimodal data, including videos and agent trajectories, are used for reinforcement learning. The training objective considers task-output correctness, evidence validity, memory read/write behavior, agreement between the model's reasoning trace and the reference reasoning trace, termination quality, and interaction cost. Through this training, VLX-VR learns what evidence to acquire, what information to retain, how to use returned Observations, and when the available evidence is sufficient to complete the task.

A possible reward decomposition is:
\begin{equation}
\begin{aligned}
R ={}& R_{\mathrm{task}} + \lambda_e R_{\mathrm{evidence}} + \lambda_m R_{\mathrm{memory}} \\
&+ \lambda_c R_{\mathrm{consistency}} - \lambda_{\mathrm{cost}} R_{\mathrm{cost}}.
\end{aligned}
\label{eq:reward}
\end{equation}
Here, $R_{\mathrm{task}}$ measures task-output quality (final-answer correctness for VideoQA), $R_{\mathrm{evidence}}$ measures whether the acquired evidence covers the task-relevant events, states, and cues identified by the reference reasoning trace, $R_{\mathrm{memory}}$ measures whether VLX-VR selects and executes useful read and write operations, $R_{\mathrm{consistency}}$ measures agreement between the model's reasoning trace and the reference reasoning trace, and $R_{\mathrm{cost}}$ penalizes unnecessary loops, tool calls, and token usage. The evidence and consistency rewards therefore measure complementary properties rather than the same behavior.

\section{Dataset}
\label{sec:dataset}

We select MINERVA as the primary evaluation dataset because it combines broad temporal coverage with explicit reasoning supervision. It includes both short and long videos, and provides detailed, human-annotated reasoning traces in addition to final answer labels. This makes it suitable for evaluating not only whether VLX-VR selects the correct answer, but also whether its reasoning trace is grounded in the relevant evidence described by the annotation~\cite{minerva}. Each question contains a video, a question, five answer choices, and a human-annotated reasoning trace. In our analysis, the events, states, temporal locations, and cues described by this trace define the reference evidence against which evidence acquisition and reasoning consistency are interpreted. The questions require at least two reasoning skills and cover sports, tutorials, daily activities, travel, and other domains. Video-MME also evaluates video understanding across domains, durations, and modality conditions~\cite{videomme}, while LongVideoBench targets long-context interleaved video-language understanding and referring reasoning~\cite{longvideobench}. Compared with these complementary benchmarks, MINERVA is particularly suitable for our analysis of answer accuracy and evidence-grounded reasoning.

The reference reasoning traces contain approximately 92 words on average. About 99.6\% contain timestamps, with approximately four timestamps per trace~\cite{minerva}. These annotations support temporal localization, reference-evidence identification, and reasoning-consistency analysis. The human accuracy reported by MINERVA is approximately 92.5\%~\cite{minerva}; we report it as human reference performance rather than as a theoretical upper bound or a result obtained under exactly the same system conditions.

We follow the original MINERVA task protocol~\cite{minerva}. The model receives a video, a question, and five candidate answers, and returns a final choice together with a VLX-VR reasoning trace for analysis. The main metric is multiple-choice accuracy. VLX-VR results are taken from the current evaluation report. Seed2.1 Pro, Seed2.1 Turbo, Gemini 3.5 Flash, and Gemini 3.1 Pro comparison numbers are linked to the official Seed2.1 page~\cite{seed21web}, while the remaining comparison numbers come from the public MINERVA evaluation~\cite{minerva}.

\section{Experimental Results}
\label{sec:experiments}

This evaluation uses the agentic-aware VLX-VR model obtained from the training procedure described above. VLX-VR operates within the framework-defined Think--Memory--Observation loop to acquire relevant multimodal evidence before answering questions from the MINERVA dataset. During Memory, VLX-VR directly invokes the provided \texttt{read\_memory} or \texttt{write\_memory} operation.

\subsection{Overall Accuracy}
The primary metric is multiple-choice accuracy:
\begin{equation}
\mathrm{Accuracy} = \frac{N_{\mathrm{correct}}}{N_{\mathrm{questions}}}.
\label{eq:accuracy}
\end{equation}
VLX-VR obtains 78.79\% overall accuracy on MINERVA. The complete comparison with other models is reported in Table~\ref{tab:comparison}.

\begin{table}[t]
  \caption{Comparison on MINERVA. Model rows are ordered by accuracy; human and random reference levels are listed separately. Seed2.1 and Gemini 3.x results are from the Seed2.1 page~\cite{seed21web}; other external results are from MINERVA~\cite{minerva}.}
  \label{tab:comparison}
  \centering
  \scriptsize
  \resizebox{\linewidth}{!}{%
  \begin{tabular}{@{}lrrl@{}}
    \toprule
    Model & Accuracy (\%) & \shortstack{$\Delta$ vs. VLX-VR\\(pp)} & Source \\
    \midrule
    \textbf{VLX-VR} & \textbf{78.79} & -- & This work \\
    Seed2.1 Pro & 70.70 & -8.09 & \cite{seed21web} \\
    Gemini 3.5 Flash & 68.60 & -10.19 & \cite{seed21web} \\
    Gemini 2.5 Pro Thinking & 66.20 & -12.59 & \cite{minerva} \\
    Seed2.1 Turbo & 65.90 & -12.89 & \cite{seed21web} \\
    Gemini 3.1 Pro & 63.50 & -15.29 & \cite{seed21web} \\
    Gemini 2.5 Flash Thinking & 57.30 & -21.49 & \cite{minerva} \\
    GPT-4.1 & 53.99 & -24.80 & \cite{minerva} \\
    GPT-4o & 45.54 & -33.25 & \cite{minerva} \\
    OpenAI o1 & 43.48 & -35.31 & \cite{minerva} \\
    VideoLLaMA3 & 35.91 & -42.88 & \cite{minerva} \\
    InternVideo2.5 & 35.18 & -43.61 & \cite{minerva} \\
    Qwen2.5-VL & 35.05 & -43.74 & \cite{minerva} \\
    Claude 3.5 Sonnet v2 & 31.28 & -47.51 & \cite{minerva} \\
    \midrule
    Human & 92.54 & +13.75 & \cite{minerva} \\
    Random & 20.00 & -58.79 & \cite{minerva} \\
    \bottomrule
  \end{tabular}}
\end{table}

VLX-VR is higher than the public comparison results in the current aggregate table, but remains below the human reference level. This suggests strong performance on MINERVA's multi-step, multi-skill video reasoning task, but does not establish universal superiority across video understanding tasks.

\subsection{Performance across Video Durations}
Table~\ref{tab:duration_comparison} evaluates long-video performance using MINERVA's three duration groups and public baseline results~\cite{minerva}.
\begin{table}[t]
  \caption{Accuracy across three video-duration groups.}
  \label{tab:duration_comparison}
  \centering
  \scriptsize
  \resizebox{\linewidth}{!}{%
  \begin{tabular}{@{}lrrr@{}}
    \toprule
    Model & Below 5 min & 5--15 min & Above 15 min \\
    \midrule
    \textbf{VLX-VR} & \textbf{76.70} & \textbf{78.73} & \textbf{80.92} \\
    Gemini 2.5 Pro Thinking & 68.87 & 66.84 & 57.97 \\
    GPT-4.1 & 58.84 & 54.79 & 47.25 \\
    OpenAI o1 & 48.28 & 41.45 & 40.38 \\
    Claude 3.5 Sonnet v2 & 40.90 & 33.68 & 28.30 \\
    \bottomrule
  \end{tabular}}
\end{table}

\begin{figure}[t]
  \centering
  \includegraphics[width=\linewidth]{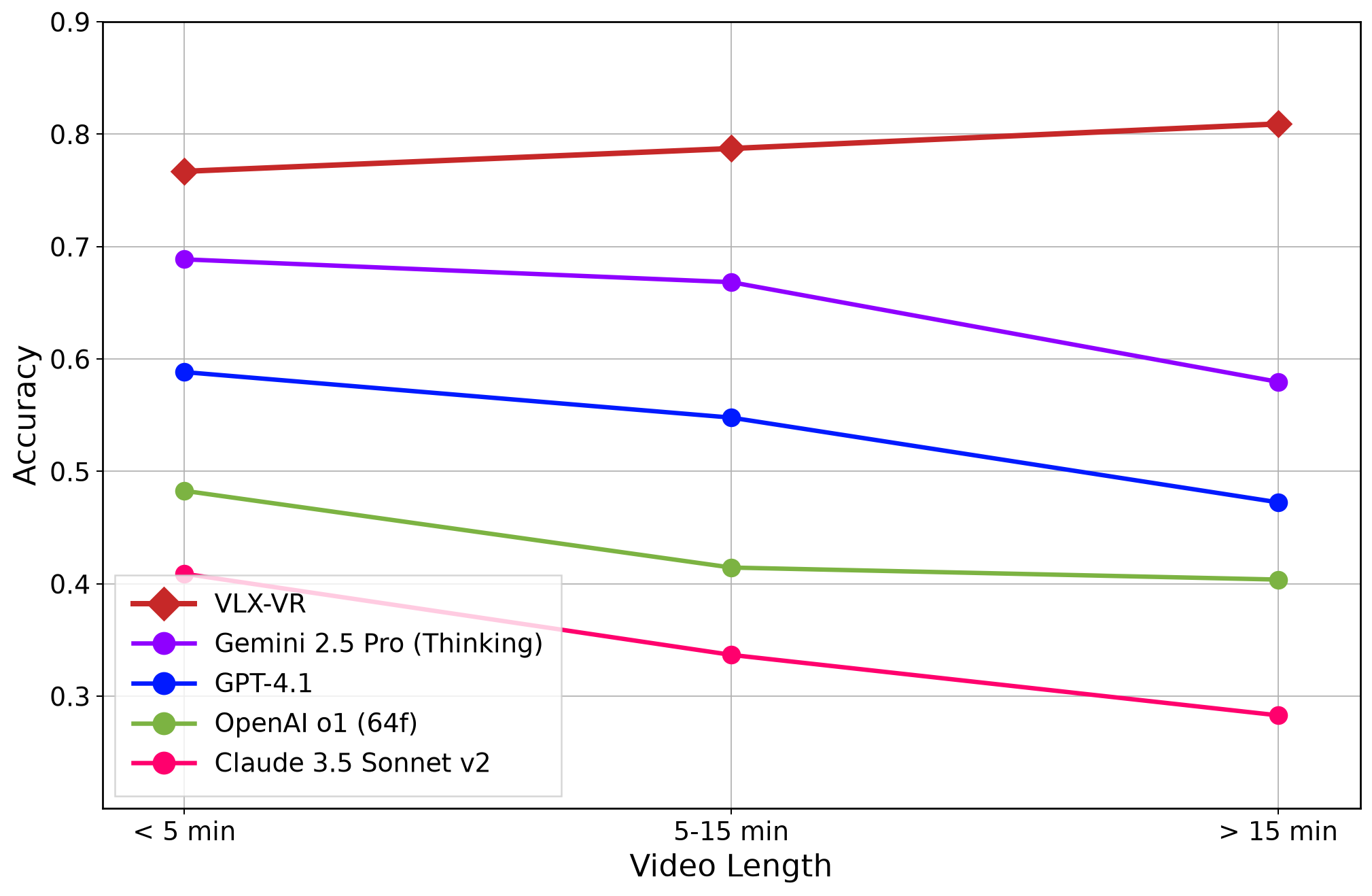}
  \caption{Accuracy comparison across video-duration groups. VLX-VR obtains 76.70\%, 78.73\%, and 80.92\% in the three groups. The current result does not show a monotonic decrease with duration.}
  \label{fig:duration}
\end{figure}

Figure~\ref{fig:duration} visualizes the three-group comparison. VLX-VR increases from 76.70\% below 5 minutes to 78.73\% at 5--15 minutes and 80.92\% above 15 minutes, whereas all four public baselines decline as video duration increases. We quantify cross-duration stability with the equally weighted population variance of accuracy across duration groups:
\begin{equation}
\mathrm{CDAV} = \frac{1}{n}\sum_{i=1}^{n}(\mathrm{Acc}_i - \overline{\mathrm{Acc}})^2.
\label{eq:cdav}
\end{equation}
Here, $\mathrm{Acc}_i$ is measured in percentage points, so CDAV is reported in squared percentage points ($\mathrm{pp}^2$). Table~\ref{tab:cdav_comparison} applies this definition to the three groups in Table~\ref{tab:duration_comparison}, with equal weighting across duration groups and statistics computed from the displayed rounded accuracies.

\begin{table}[H]
  \caption{Cross-duration performance statistics under MINERVA's original duration grouping. Lower CDAV indicates less variation across duration groups, but does not by itself imply higher accuracy.}
  \label{tab:cdav_comparison}
  \centering
  \scriptsize
  \resizebox{\linewidth}{!}{%
  \begin{tabular}{@{}lrrrr@{}}
    \toprule
    Model & Mean accuracy & CDAV & Std. & Range \\
    & (\%) & (pp$^2$) & (pp) & (pp) \\
    \midrule
    \textbf{VLX-VR} & \textbf{78.78} & \textbf{2.97} & \textbf{1.72} & \textbf{4.22} \\
    Gemini 2.5 Pro Thinking & 64.56 & 22.40 & 4.73 & 10.90 \\
    GPT-4.1 & 53.63 & 23.06 & 4.80 & 11.59 \\
    OpenAI o1 & 43.37 & 12.24 & 3.50 & 7.90 \\
    Claude 3.5 Sonnet v2 & 34.29 & 26.65 & 5.16 & 12.60 \\
    \bottomrule
  \end{tabular}}
\end{table}

Under MINERVA's original duration grouping, VLX-VR has a mean accuracy of 78.78\%, a CDAV of $2.97~\mathrm{pp}^2$, a standard deviation of 1.72~pp, and a range of 4.22~pp. It has the lowest CDAV among the five models while also achieving the highest mean accuracy. The other four models have CDAV values of 22.40, 23.06, 12.24, and 26.65~$\mathrm{pp}^2$, respectively. Thus, under the original MINERVA grouping, VLX-VR maintains stable performance as duration increases; however, the aggregated Above 15 min group does not reveal how performance varies within longer videos.

\subsection{Performance Analysis on Different Durations}
The analysis under MINERVA's original grouping shows the unusual pattern that VLX-VR performs better as video duration increases, reaching 80.92\% in the aggregated Above 15 min group. To determine whether this improvement is uniform across longer videos, we divide that group into 15--30 min and Above 30 min intervals while retaining the original Below 5 min and 5--15 min groups. Table~\ref{tab:four_duration} reports the resulting refined analysis.

\begin{table}[H]
  \caption{Refined duration analysis with videos longer than 15 minutes split at 30 minutes.}
  \label{tab:four_duration}
  \centering
  \begin{tabular}{@{}lc@{}}
    \toprule
    Video duration & Accuracy (\%) \\
    \midrule
    0--5 min & 76.70 \\
    5--15 min & 78.73 \\
    15--30 min & 83.40 \\
    Above 30 min & 74.75 \\
    \bottomrule
  \end{tabular}
\end{table}

The highest accuracy occurs in the 15--30 min interval at 83.40\%, whereas accuracy decreases to 74.75\% above 30 minutes. The apparent improvement in the original Above 15 min result is therefore driven primarily by the strong 15--30 min interval rather than by a uniform increase across all longer videos. We next compare the cross-duration statistics under the original and refined groupings in Table~\ref{tab:cdav}.

\begin{table}[H]
  \caption{Cross-duration performance statistics for VLX-VR under the original MINERVA and refined duration groupings.}
  \label{tab:cdav}
  \centering
  \scriptsize
  \resizebox{\linewidth}{!}{%
  \begin{tabular}{@{}lrrrrr@{}}
    \toprule
    Grouping & Groups & Mean accuracy & CDAV & Std. & Range \\
    & & (\%) & (pp$^2$) & (pp) & (pp) \\
    \midrule
    Original MINERVA & 3 & 78.78 & 2.97 & 1.72 & 4.22 \\
    Refined & 4 & 78.40 & 10.33 & 3.21 & 8.65 \\
    \bottomrule
  \end{tabular}}
\end{table}

Both analyses assign equal weight to each duration group, and the statistics are computed from the rounded accuracies reported in Tables~\ref{tab:duration_comparison} and~\ref{tab:four_duration}. Splitting the long-video group changes the equal-weight mean only slightly, from 78.78\% to 78.40\%, but increases CDAV from 2.97 to 10.33~$\mathrm{pp}^2$, standard deviation from 1.72 to 3.21~pp, and range from 4.22 to 8.65~pp. The higher variation under the refined grouping shows that the original MINERVA grouping smooths over the 15--30 min peak and the subsequent decline above 30 minutes. The strong 15--30 min result suggests that the learned evidence-acquisition behavior of the agentic-aware VLX-VR model broadens the duration range over which it can effectively analyze video. Although performance declines beyond 30 minutes, the refined-group accuracies remain within 74.75--83.40\%, indicating broadly stable performance rather than a collapse on longer videos.

\subsection{Performance across Reasoning Skills}
Table~\ref{tab:skills} reports the accuracy distribution across reasoning skills.
\begin{table}[H]
  \caption{VLX-VR accuracy by reasoning skill.}
  \label{tab:skills}
  \centering
  \begin{tabular}{@{}lr@{}}
    \toprule
    Skill & Accuracy (\%) \\
    \midrule
    Counting & 66.67 \\
    Object recognition & 82.01 \\
    Event occurrence & 82.81 \\
    Listening & 75.19 \\
    Temporal reasoning & 85.95 \\
    Reading & 90.76 \\
    Numerical reasoning & 85.71 \\
    Spatial perception & 73.47 \\
    Cause and effect & 72.73 \\
    Counterfactual reasoning & 74.19 \\
    Situational awareness & 90.32 \\
    Goal reasoning & 76.92 \\
    State changes & 69.23 \\
    \bottomrule
  \end{tabular}
\end{table}
VLX-VR is strongest on reading, situational awareness, temporal reasoning, and numerical reasoning. Event occurrence and object recognition are also above the overall accuracy. Counting, state changes, cause and effect, and spatial perception are below the overall level, suggesting the need for more reliable object tracking, state modeling, local evidence retrieval, and causal judgment. The skill results describe the capability distribution and do not by themselves prove that a particular module caused a given difference.

\subsection{Analysis of Reference-trace Agreement}
On correctly answered samples, 96.20\% of the VLX-VR reasoning traces are judged consistent with the MINERVA reference reasoning traces and the evidence described by them. Across the full evaluation set, the proportion of samples estimated to have both a correct answer and a VLX-VR reasoning trace consistent with the reference reasoning trace is $78.79\% \times 96.20\% \approx 75.80\%$. This joint criterion is stricter than answer accuracy alone. Nevertheless, the estimated joint rate is approximately 5.10 percentage points higher than the published answer-only accuracy of Doubao Seed2.1 Pro (70.70\%)~\cite{seed21web}.

This comparison is descriptive because the two results use different success criteria: VLX-VR is evaluated by the joint requirement of answer correctness and agreement between its reasoning trace, the MINERVA reference reasoning trace, and the evidence described by that trace, whereas Seed2.1 Pro reports answer accuracy only. The reported agreement is therefore an evaluation of evidence-grounded consistency with the annotation, not a direct measurement of the raw evidence itself or a guarantee that the trace faithfully reveals the model's internal decision process~\cite{cotfaithfulness}.

\section{Case Studies}
\label{sec:case_studies}

To connect the aggregate results above with concrete model behavior, this section presents detailed positive and negative cases covering temporal reasoning, numerical reasoning, situational awareness, counting, and state changes. The videos, questions, answer options, reference answers, and reference reasoning traces are from MINERVA~\cite{minerva}; the VLX-VR answers, reasoning traces, and case analyses are from our evaluation. The cases distinguish the evidence observed by VLX-VR from the reasoning trace that organizes and interprets that evidence, making the reference answer, the VLX-VR answer, and the resulting strengths or limitations explicit.

\subsection{Correct case 1: Temporal reasoning}
\textbf{Video:} \href{https://www.youtube.com/watch?v=8rdIuLhEQso}{resource-169422}, approximately 257 seconds.

\textbf{Question:} Who is the second person to speak in the video?

\textbf{Options:} A narrator; The brunette woman; It is a silent film; The blonde woman; The brunette woman's fiancee.

\textbf{Reference answer:} The blonde woman.

\textbf{Reference reasoning trace:} I watch the video until I see the brunette woman speak at 00:41. When I see the blonde woman speak at 00:47, then, I know that she is the second woman to speak in the video.

\textbf{VLX-VR answer:} The blonde woman; correct.

\textbf{VLX-VR reasoning trace:} VLX-VR first reviewed the full video and confirmed that only two women appear. It then divided the 40--48 second interval into multiple observation windows, identifying the brunette woman as the first speaker around 41 seconds and the blonde woman as the responder around 45.6--48 seconds based on lip motion and shot reversals. It further used later alternating dialogue to cross-check the speaking order.

\begin{figure*}[t]
  \centering
  \includegraphics[width=\textwidth]{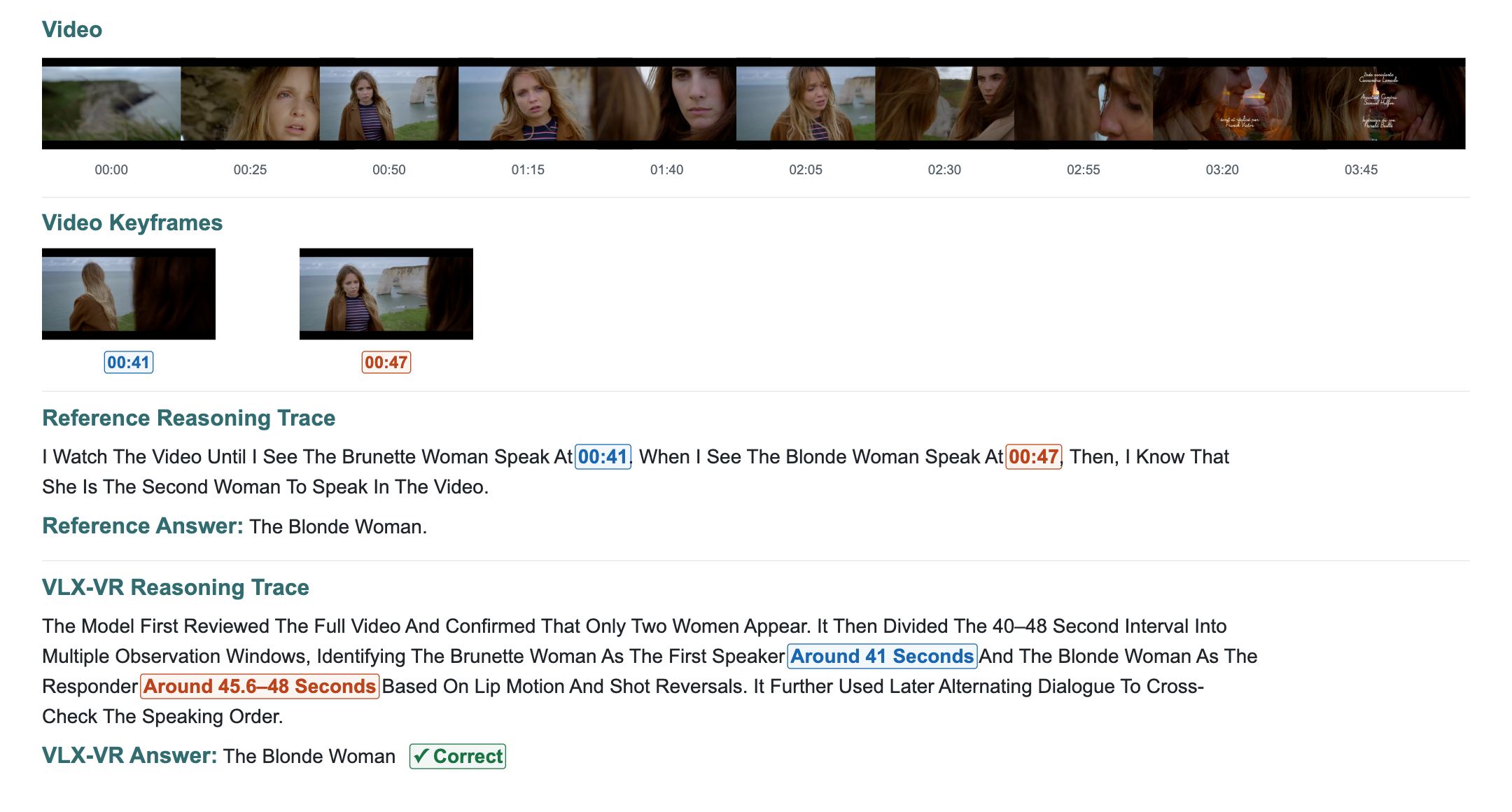}
  \caption{Visualization of the temporal-reasoning case for resource-169422. The full-video overview supports coarse temporal localization, while the keyframes at 00:41 and 00:47 identify the first and second speakers. The lower panels align the MINERVA reference reasoning trace with the VLX-VR reasoning trace.}
  \label{fig:case-temporal}
\end{figure*}

\textbf{Analysis:} The question requires event ordering rather than simple person identification. The VLX-VR reasoning trace contains a chain of first-speaking timestamp, speaker identity, and subsequent alternating dialogue, connecting local visual observations with temporal order. Within the framework-defined Think--Memory--Observation loop, VLX-VR uses the first Observation to locate the first utterance, the second to verify the second speaker, and later observations to test whether the speaking sequence remains coherent. Figure~\ref{fig:case-temporal} summarizes the full-video scan, the decisive keyframes, and the supporting evidence described in the VLX-VR reasoning trace.

\subsection{Correct case 2: Numerical reasoning}
\textbf{Video:} \href{https://www.youtube.com/watch?v=iQGTWM1jkgE}{resource-169908}, approximately 5,686 seconds.

\textbf{Question:} When orange first takes control of the four territories of Australia, what is the average value per territory that green controls?

\textbf{Options:} 2.9; 2.8; 3.2; 2.7; 2.3.

\textbf{Reference answer:} 2.7.

\textbf{Reference reasoning trace:} I identified Australia on the map as the continent in the bottom right section of the frame. At 12:11, I saw the leftmost territory of mainland Australia attack the chain of islands west of New Zealand. I saw that the attack was successful, as the chain of islands was now colored orange and had orange ``10'' on it. At this point, this territory, New Zealand, and the left and right mainland territories of Australia, all glowed orange. I understood this to mean that the orange player now controlled the entire continent. I observed the rest of the map and looked for green territories. I counted 10 territories colored green, each with their own cloud-shaped green icon on them that had a number inside. I understood from my knowledge of Risk that this corresponds to the number of troops in that territory. From left to right, top to bottom, I recorded the number of troops present in each of the territories: 1, 1, 4, 4, 3, 3, 6, 3, 1, 1. I added these together to get the total value: 1+1+4+4+3+3+6+3+1+1=27. To find the average value per territory, I divided 27/10=2.7.

\textbf{VLX-VR answer:} 2.7; correct.

\textbf{VLX-VR reasoning trace:} VLX-VR first used a coarse full-video scan to locate the interval around 695--775 seconds. It then used fine-grained observation to confirm that western Australia changed from black to orange around 712--713 seconds. It read the green player's panel, verified 10 green territories and a total value of 27, and computed 27 / 10 = 2.7.

\textbf{Analysis:} This case shows a long-video chain of event trigger, state reading, and numerical computation. Rather than guessing from a local frame, the model first localized the state change in which orange completed control of Australia and then read the global panel for green. During Memory, VLX-VR can use \texttt{write\_memory} to retain the time of the territory change and the ``green 27 / 10'' state, while \texttt{read\_memory} retrieves this evidence for the final arithmetic step. The retrieved evidence is returned as an Observation.

\subsection{Correct case 3: Situational awareness}
\textbf{Video:} \href{https://www.youtube.com/watch?v=cxHtplim5Ic}{resource-169911}, approximately 338 seconds.

\textbf{Question:} Before the man kisses the other man, what does he do before this?

\textbf{Options:} He looks down; He walks away; He gestures around; He looks up; He looks around.

\textbf{Reference answer:} He looks around.

\textbf{Reference reasoning trace:} I watched the video to find the moment when the man kisses the other man. This occurs from 03:06--03:10. The most recent scene before this which features both men runs from 02:53--03:01. During this scene, the man looks down and around the two of them, up and down the stairwell where they are standing. Thus, he looks around before he kisses the other man.

\textbf{VLX-VR answer:} He looks around; correct.

\textbf{VLX-VR reasoning trace:} VLX-VR localized a continuous stairwell segment and recorded the actions from approximately 173--181 seconds. The man first looked at the other person, then turned his gaze downward, upward, and toward the upper-left area along the railing before turning back and approaching the other man. VLX-VR ruled out ``walks away'' and the isolated ``looks down/looks up'' options and selected ``looks around.''

\textbf{Analysis:} The decisive evidence is not a single action but a sequence of gaze directions forming a situation. The model interpreted looking down and looking up as instances of the higher-level action ``looking around,'' while using the surrounding interaction to disambiguate the options. This suggests that situational-awareness questions can benefit from repeated Observations that aggregate local actions before Think performs semantic abstraction.

\subsection{Incorrect case 1: Counting under occlusion and uncertainty}
\textbf{Video:} \href{https://www.youtube.com/watch?v=IEP7f0uWURs}{resource-169417}, approximately 221 seconds.

\textbf{Question:} How many women appear in the first row of the audience?

\textbf{Options:} 6; 4; 5; 3; 2.

\textbf{Reference answer:} 5.

\textbf{Reference reasoning trace:} I watched the video to identify the times that the camera focuses on the audience. This happens at 01:04, 01:08, 02:18, and 03:39. Each instance features the same shot of the audience. From there, I counted how many women appeared sitting in the first row, which is 5.

\textbf{VLX-VR answer:} 4; incorrect.

\textbf{VLX-VR reasoning trace:} VLX-VR reviewed multiple audience windows and concluded that the first row contained seven seats: four people whose gender it could confirm as female, two males, and one person whose torso and legs were visible but whose face was cropped out. It adopted a conservative policy, excluded the uncertain person, and returned 4.

\textbf{Error cause:} The model did not fail because it lacked all relevant evidence. Instead, it applied a ``count only explicitly confirmed instances'' policy under occlusion and cropping. This disagreed with the benchmark annotation, which counted five women, producing an undercount of one. The failure lies in the definition of the target set and the treatment of uncertain instances, rather than in arithmetic.

\textbf{Model limitation:} Large models remain sensitive to occlusion, cropping, row assignment, and identity uncertainty in multi-object counting. The model can identify local attributes, but it may fail to merge observations across shots into the same object set as the benchmark. When visual evidence is incomplete, both conservative exclusion and unsupported inference can cause errors. VLX-VR may request additional observations to reduce uncertainty, but the framework-defined loop in which the agentic-aware VLX-VR model operates cannot guarantee recovery of an attribute that is absent from the available visual evidence.

\subsection{Incorrect case 2: State changes and temporal-chain tracking}
\textbf{Video:} \href{https://www.youtube.com/watch?v=9120Php3Kh4}{resource-169421}, approximately 527 seconds.

\textbf{Question:} What change of color occurs to the letters added to the word ``ZING'' by Chris Cree?

\textbf{Options:} They turn from red to yellow; They turn from yellow to red; They turn from green to red; They turn from yellow to green; They turn from red to green.

\textbf{Reference answer:} They turn from yellow to green.

\textbf{Reference reasoning trace:} I watched for the section of video dedicated to Chris Cree and his addition to the word ``ZING'', which I found from 03:06--03:38. Since the letters ``ANNUALI'' are added to the ``ZING,'' ``ANNUALI'' are then the letters whose color I pay attention to. Initially, the letters ``ANNUALI'' are displayed in yellow, but this color changes to green at 03:05 after they are played on the board. Therefore, the letters added to the word ``ZING'' by Chris Cree change color by turning from yellow to green.

\textbf{VLX-VR answer:} They turn from yellow to red; incorrect.

\textbf{VLX-VR reasoning trace:} VLX-VR localized the Chris Cree segment around 03:06--03:38 and identified the added letters as ``ANNUALI.'' It observed that the letters turned green after being placed on the board around 185 seconds, but later observed them turn red around 215--218 seconds. It treated the later transition as the answer and described the change as ``green to red.'' The final option marker was also inconsistent with the textual conclusion: it corresponded to ``yellow to red,'' while the explanation described ``green to red.''

\textbf{Error cause:} The question asks for the transition from the initial state to the state immediately after the letters are placed. The model continued tracking a later ``unacceptable'' state and allowed this salient event to overwrite the earlier transition required by the question. It detected multiple real changes but failed to select the correct pair of states under the question's temporal boundary. The inconsistency between the explanation and the final option further indicates state drift between reasoning and answer decoding.

\textbf{Model limitation:} Large models can be attracted to later salient events in state-change questions and may fail to maintain a complete temporal chain involving object identity, initial state, target state, and event boundary. Even when the model recognizes the letters and multiple color states, it can answer incorrectly if it does not lock onto the first valid transition required by the question. During Memory, VLX-VR should use \texttt{write\_memory} to store object identity and time-ordered state records explicitly, while \texttt{read\_memory} should return structured ``before--trigger--after'' evidence as an Observation. Merely increasing the number of observations does not automatically solve event-boundary selection.

\subsection{Case-study summary: separate observation from evidence decisions}
The three correct cases show that VLX-VR can form coherent evidence chains for temporal order, numerical computation, and situational action understanding. The two incorrect cases show that VLX-VR may observe relevant frames yet fail in uncertain-object counting, selecting among multiple state transitions, or mapping its reasoning to the final option. Case analysis should therefore evaluate more than whether VLX-VR ``saw'' the relevant content. It should separately measure whether VLX-VR found the correct time interval, maintained the correct object identity, selected the state transition requested by the question, and kept the VLX-VR answer consistent with its reasoning trace. These are the behaviors that process rewards and structured agent trajectories should reinforce and validate in VLX-VR.

\section{Conclusion}
\label{sec:conclusion}

We presented VLX-VR, an agentic-aware video reasoning model trained to operate within a video reasoning framework organized as a Think--Memory--Observation loop. Within this framework-defined loop, VLX-VR identifies what evidence is needed, selects and executes the \texttt{read\_memory} or \texttt{write\_memory} operation provided by Memory, integrates the returned Observation, and decides whether to continue gathering evidence or produce the task output. Multimodal memory serves as an external and inspectable reasoning state for video, audio, supporting evidence, Observations, and intermediate states. Consequently, evidence acquisition and memory use become model behaviors that can be recorded, analyzed, and trained rather than remaining implicit components of an external inference pipeline.

On MINERVA, VLX-VR achieves state-of-the-art performance among the models included in our comparison, with 78.79\% overall accuracy. Across MINERVA's three duration groups, it obtains accuracies of 76.70\%, 78.73\%, and 80.92\%, together with the lowest CDAV under the original MINERVA grouping among the five models analyzed. The finer-grained duration analysis shows that the apparent improvement for longer videos is driven primarily by the 83.40\% result in the 15--30 min interval; accuracy decreases to 74.75\% above 30 minutes rather than increasing monotonically with duration. On correctly answered samples, 96.20\% of VLX-VR's reasoning traces are consistent with the MINERVA reference reasoning traces and the evidence described by them, and approximately 75.80\% of all evaluated samples satisfy both answer correctness and this evidence-grounded trace criterion.

These results support the effectiveness of VLX-VR across varied video durations and reasoning skills, while also defining clear limitations. Performance remains weaker for counting, state changes, cause and effect, and spatial perception, and evidence-grounded agreement with the reference reasoning trace indicates consistency with annotated reasoning and observed evidence rather than guaranteed faithfulness to the model's internal decision process. Separating the causal contributions of the framework-defined loop, direct multimodal-memory access, and reinforcement-learning training will require matched ablations and remains an important direction for future evaluation.

\newpage
{
    \small
    \bibliographystyle{ieeenat_fullname}
    \bibliography{main}
}

\end{document}